\documentclass{article}
\usepackage{spconf,amsmath,graphicx}
\usepackage{xcolor, bm, multirow}
\usepackage{amssymb}
\usepackage{subcaption}
\usepackage{xcolor}
\usepackage[colorlinks=true, allcolors=blue]{hyperref}
\usepackage{orcidlink}
\usepackage{etoolbox}
\usepackage{balance}

\usepackage{color,soul}

\title{Data Leakage in Patch-Based Hyperspectral Image Classification: Quantifying the Impact of Spatial Overlap}

 \name{Mohammed Q. Alkhatib $^{\orcidlink{0000-0003-4812-614X}}$ 
 }

\address{College of Engineering and IT, University of Dubai, Dubai, 14143, UAE \\mqalkhatib@ieee.org}

\begin{document}

\maketitle

\begin{abstract}
Patch-based learning improves hyperspectral image (HSI) classification by exploiting local spectral-spatial information, but random train-test sampling from the same image can cause spatial patch overlap, leading to data leakage and optimistic performance estimates. This paper investigates same-class train-test spatial overlap in patch-based HSI classification using two measures: overlap percentage ($OP$), which quantifies the global amount of overlapped testing patch pixels, and average overlap ratio ($AOR$), which measures the local severity among affected testing patches. Experiments on the Pavia University dataset compare random and non-random spatial sampling using SVM, MLP, 2D-CNN, 3D-CNN, ViT, and MorpMamba. The results show that deep patch-based models achieve high accuracy under random sampling, with 3D-CNN reaching 96.17\% Overall Accuracy (OA), but drop substantially under non-random spatial sampling, where 3D-CNN decreases to 55.20\% and ViT and 2D-CNN drop by 40.71 and 38.81 percentage points (PP), respectively. Patch-size analysis further shows that increasing the patch size from $5 \times 5$ to $19 \times 19$ raises the random-sampling overlap percentage from 23.28\% to 77.02\%. These findings demonstrate that random patch-based evaluation can substantially inflate classification performance, especially for models that strongly exploit spatial context. The code associated with this paper is available at: \url{https://github.com/mqalkhatib/Data_Leakage_in_HSI_Classification}.

\end{abstract}
\begin{keywords}
HSI classification, Data Leakage, Spatial Overlap, Deep Learning, Patch-Based Learning
\end{keywords}

\section{Introduction}
\vspace{-1em}
Hyperspectral image (HSI) classification assigns semantic labels to pixels using their spectral responses across many contiguous bands. Compared with Optical imagery, HSI captures richer material-specific information, which improves the discrimination of visually similar land-cover classes. This makes HSI classification important for fine-grained scene understanding, particularly when different materials exhibit similar spatial appearance but distinct spectral signatures. However, the task remains challenging due to high dimensionality, spectral redundancy, spatial variability within the same class, and the limited availability of labeled samples.

Traditional methods, such as Support Vector Machines (SVMs) \cite{melgani2004classification} and Multilayer Perceptrons (MLPs) \cite{plaza2008parallel}, have been widely used for HSI classification, but they often depend mainly on spectral features and have limited ability to exploit spatial context. Deep learning methods address this limitation by learning spectral-spatial representations directly from HSI data \cite{Alkhatib26042026}. Convolutional neural network (CNN)-based models have progressed from spectral 1D-CNNs \cite{hu2015deep} to spatial 2D-CNNs \cite{makantasis2015deep}, joint spectral-spatial 3D-CNNs \cite{hamida20183}, and hybrid 2D-3D CNNs \cite{roy2019hybridsn}. More recently, transformer- and Mamba-based models have been introduced to capture long-range dependencies and improve spectral-spatial modeling \cite{alkhatib2024hsiformer, ahmad2025spatial}.

Despite these advances, many HSI classifiers still rely on patch-based learning, where a spatial neighborhood around each labeled pixel is used as input. Although this improves spectral-spatial representation, it can introduce data leakage when training and testing samples are randomly selected from the same image. In such cases, their patches may spatially overlap, causing shared pixels to appear in both sets, weakening train-test independence and producing overly optimistic performance estimates \cite{feng2023information, zhan2026rethinking}. This leakage becomes more severe with larger patch sizes and higher training ratios, and is further amplified by spatial autocorrelation, where neighboring pixels often have similar spectra \cite{zhan2026rethinking}. Although spatially disjoint sampling can reduce this issue \cite{ahmad2022disjoint}, it may further reduce the already limited number of labeled training samples.

To better understand and quantify data leakage,
this paper investigates the relationship between same-class
train-test spatial overlap and classification performance across
six classifiers and two sampling protocols. Unlike prior studies
that mainly identify or mitigate leakage, this study jointly
quantifies its global extent and local severity using overlap
percentage and average overlap ratio. SVM, MLP, 2D-CNN,
3D-CNN, Vision Transformer (ViT), and MorpMamba are evaluated
under random and non-random spatial sampling to assess model
sensitivity and quantify the resulting accuracy inflation.

\section{Related Work}
\vspace{-1em}
\subsection{Patch-Based Learning in HSI}
\vspace{-0.5em}
Patch-based learning is widely used to incorporate spatial
context by extracting a local neighborhood around each labeled
pixel and using it as the model input. This strategy allows
classifiers to exploit local spectral-spatial correlations and has
become common in CNN, Transformer, and Mamba-based HSI
classification frameworks
\cite{makantasis2015deep, hamida20183,
roy2019hybridsn,alkhatib2024hsiformer,ahmad2025spatial}.

However, patch extraction introduces dependencies between
samples from the same scene. Since neighboring patches may
share many pixels, training and testing samples can become
highly correlated even when their center pixels are different
\cite{feng2023information,zhan2026rethinking}.
This dependency is mainly controlled by the patch size and
sampling strategy. Larger patches increase spatial coverage and
therefore raise the probability of train-test overlap, while random
sampling from a single image can place training and testing pixels
close to each other due to spatial continuity
\cite{feng2023information,zhan2026rethinking}.

\subsection{Data Leakage in Remote Sensing and Computer Vision}
\vspace{-0.5em}
Data leakage occurs when testing information becomes indirectly
available during training, leading to overly optimistic performance
estimates. In patch-based HSI classification, random sampling from
the same image may place training and testing samples in spatially
correlated regions, causing their patches to overlap and share pixels
\cite{feng2023information,zhan2026rethinking}. Spatially disjoint,
iterative, and augmentation-based sampling strategies have been
proposed to reduce this dependence
\cite{ahmad2022disjoint,chang2023iterative,feng2023information},
but they may reduce the number of usable labeled samples and
primarily focus on mitigation.

Comparatively little attention has been given to
jointly quantifying the global extent and local severity of
same-class train-test patch overlap and relating these characteristics to performance degradation across different classifier families under matched experimental conditions. This work addresses this gap using two complementary overlap measures evaluated across traditional machine-learning, CNN-based, Transformer-based, and Mamba-based classifiers under random and leakage-reduced spatial sampling protocols.

\section{Methodology}
\vspace{-1em}
\subsection{Patch-Based HSI Classification}
\begin{figure}[!t]
\centering
\includegraphics[clip=true, trim = 40 5 20 5,width= 0.99\linewidth]{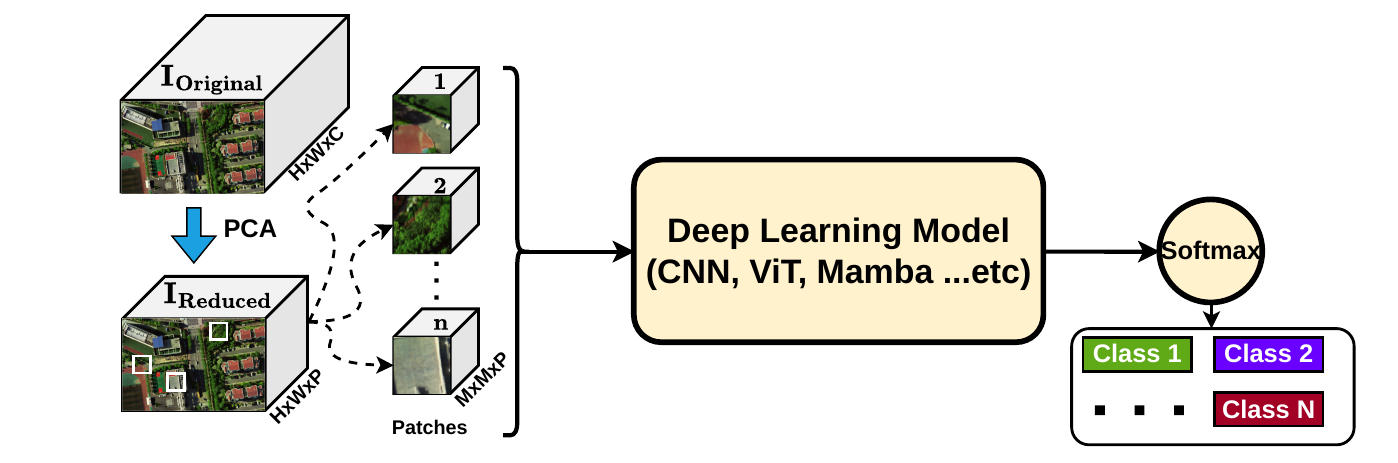}
\vspace{-1em}
\caption{General Patch-based hyperspectral image classification framework using PCA and deep learning models.}
 \label{fig:model}
\end{figure}

\vspace{-0.5em}
The general architecture of most deep learning frameworks for hyperspectral image classification is illustrated in Fig.~\ref{fig:model}. The process begins with a hyperspectral image $\mathbf{I}_{\text{Original}} \in \mathbb{R}^{H \times W \times C}$. To reduce spectral redundancy and computational cost, Principal Component Analysis (PCA) is applied, resulting in a compressed image $\mathbf{I}_{\text{Reduced}} \in \mathbb{R}^{H \times W \times P}$, where $P \ll C$. Spatial patches of size $M \times M \times P$ are then extracted and used as inputs to a deep learning model, such as CNN, ViT, or Mamba-based architectures. The extracted features are subsequently passed through a classification head with a softmax activation function to produce the final class predictions.

\subsection{Quantifying Train-Test Spatial Overlap}
\vspace{-0.5em}
To measure the amount of same-class spatial overlap between
training and testing patches, two simple measures are used.
Let $T_{\mathrm{tr}}$ and $T_{\mathrm{te}}$ denote the training
and testing center pixels, respectively. For any pixel $p$, the
patch extracted around it is denoted by $\Omega_M(p)$, where
$M$ is the patch size.

For a class $c$, the spatial coverage of its
training patches is defined as
\begin{equation}
\Omega_{\mathrm{tr}}^{(c)}
=
\bigcup_{\substack{p\in T_{\mathrm{tr}}\\c_p=c}}
\Omega_M(p),
\end{equation}
where $c_p$ denotes the class label of the training center
pixel $p$.

A testing patch is considered overlapped when it shares pixels
with the training patch coverage of the same class. This
same-class condition is used because overlap between patches
from different classes does not represent direct class-consistent
leakage.

The first measure is the overlap percentage ($OP$). It measures how many pixels inside the testing patches are also covered by
training patches from the same class. It is computed as:
\begin{equation}
OP =
\frac{
\displaystyle\sum_{q\in T_{\mathrm{te}}}
\left|
\Omega_M(q)\cap\Omega_{\mathrm{tr}}^{(c_q)}
\right|
}{
\displaystyle\sum_{q\in T_{\mathrm{te}}}
\left|\Omega_M(q)\right|
}
\times 100,
\label{eq:overlap_percentage}
\end{equation}
where $c_q$ is the class label of the testing center pixel $q$.
A higher value of $OP$ means that a larger portion of all
testing-patch pixels is also contained in same-class training
patches.

To compute the second measure, the overlap ratio of
each testing patch $q$ is first calculated as:
\begin{equation}
OR(q)=
\frac{
\left|
\Omega_M(q)\cap\Omega_{\mathrm{tr}}^{(c_q)}
\right|
}{
\left|\Omega_M(q)\right|
}.
\end{equation}

The second measure, termed the average overlap ratio
($AOR$), is then obtained by averaging $OR(q)$ over only the testing patches affected by same-class overlap:
\begin{equation}
AOR =
\frac{1}{|Q|}
\sum_{q\in Q} OR(q),
\label{eq:average_overlap_ratio}
\end{equation}
where
$Q=\{q\in T_{\mathrm{te}}:OR(q)>0\}$ is the set of testing patches with non-zero same-class overlap.

Therefore, $OP$ measures the global prevalence
of leakage across the complete testing set, whereas $AOR$
measures the conditional severity of leakage among only the
affected testing patches. Consequently, a sampling protocol
may produce a low $OP$ but a relatively high $AOR$ when only
a small number of testing patches overlap substantially with
same-class training patches.

\subsection{Non-Random Spatial Sampling Protocol}
\begin{figure}[!t]
\centering
\includegraphics[clip=true, trim = 40 5 20 5,width= 0.95\linewidth]{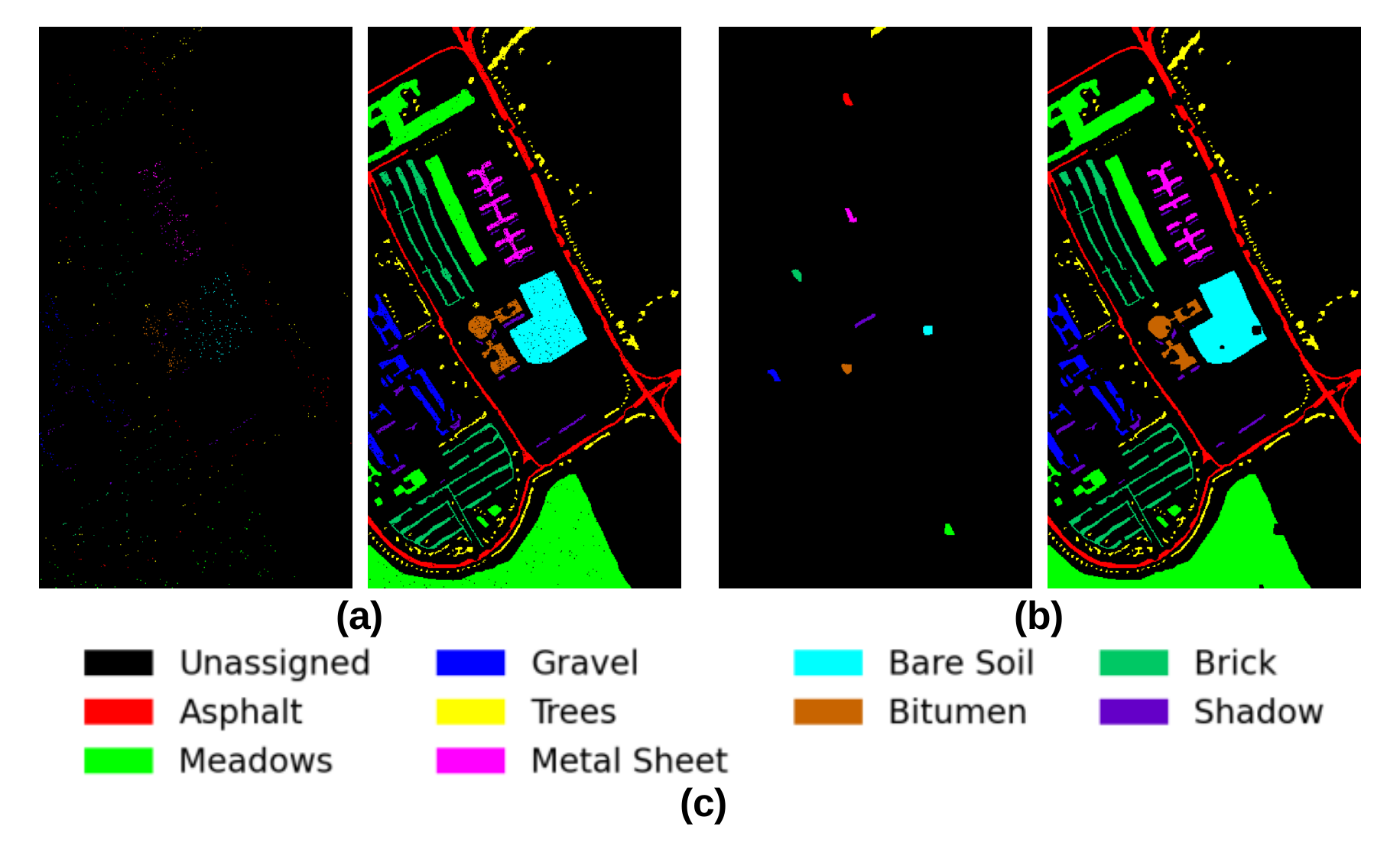}
\vspace{-1em}
\caption{Examples of the sampling settings used in this study: (a) random sampling, where training and testing patches may overlap; (b) non-random spatial sampling, where training and testing regions are selected with greater spatial separation; (c) Class Labels.}

\vspace{-1em}
 \label{fig:splits}
\end{figure}

\vspace{-0.5em}
To examine the effect of spatial overlap in patch-based HSI classification, two sampling protocols are considered, as illustrated in Fig.~\ref{fig:splits}. The first follows the conventional random strategy, where labeled pixels are divided into training, validation, and testing subsets. Since each pixel is represented by a spatial patch, nearby training and testing samples may share pixels, introducing spatial dependency and potentially inflating classification performance.

The second protocol uses non-random spatial sampling, where samples are selected with greater spatial separation. This reduces direct mixing between training and testing regions and provides a stricter evaluation setting. Since testing pixels are not explicitly excluded after patch extraction, this setting is considered leakage-reduced rather than fully leakage-free. Training samples are selected first. Validation
samples are then drawn from the remaining labeled pixels and
used only for model monitoring and selection, while all other
labeled pixels form the test set used for the final performance
evaluation.

Comparing the two protocols allows the sensitivity of each classifier to train-test spatial overlap to be evaluated. A large performance drop under non-random spatial sampling suggests that the model may have benefited from spatial dependency under random patch-based evaluation.

\subsection{Experimental Setup}
\vspace{-0.5em}
\begin{table*}[t!]
\centering
\caption{Class-wise and overall classification performance under the random patch-based evaluation setting with potential same-class train-test spatial overlap. The reported values are presented as mean$\pm$standard deviation over 10 runs.}
\vspace{-1em}
\label{Random}
\resizebox{0.85\linewidth}{!} {
\begin{tabular}{clcccccccccc}
\hline
No & Name        & Train & Val & Test   & Total  & SVM            & MLP             & 2D-CNN         & 3D-CNN         & ViT            & MorpMamba       \\ \hline
1  & Asphalt     & 94    & 18  & 6,519  & 6,631  & 62.83$\pm$1.01 & 64.51$\pm$3.11  & 83.07$\pm$8.19 & 94.28$\pm$1.95 & 89.26$\pm$3.32 & 75.67$\pm$6.86  \\
2  & Meadows     & 94    & 18  & 18,537 & 18,649 & 49.46$\pm$4.91 & 64.17$\pm$8.45  & 84.98$\pm$5.47 & 97.63$\pm$1.17 & 91.32$\pm$4.04 & 78.05$\pm$7.50  \\
3  & Gravel      & 94    & 18  & 1,987  & 2,099  & 21.40$\pm$8.34 & 69.26$\pm$12.62 & 83.04$\pm$4.28 & 91.59$\pm$3.10 & 88.94$\pm$3.86 & 74.49$\pm$16.57 \\
4  & Trees       & 94    & 18  & 2,952  & 3,064  & 95.32$\pm$1.48 & 96.14$\pm$1.50  & 96.98$\pm$1.04 & 97.74$\pm$1.11 & 97.71$\pm$1.11 & 95.45$\pm$1.64  \\
5  & Metal Sheet & 94    & 18  & 1,233  & 1,345  & 99.63$\pm$0.15 & 99.65$\pm$0.20  & 99.50$\pm$0.53 & 99.99$\pm$0.02 & 99.99$\pm$0.02 & 99.94$\pm$0.06  \\
6  & Bare Soil   & 94    & 18  & 4,917  & 5,029  & 67.18$\pm$5.16 & 64.71$\pm$11.94 & 88.51$\pm$3.45 & 96.38$\pm$1.56 & 89.44$\pm$4.82 & 88.41$\pm$6.44  \\
7  & Bitumen     & 94    & 18  & 1,218  & 1,330  & 87.40$\pm$1.55 & 87.23$\pm$5.45  & 92.13$\pm$3.65 & 97.04$\pm$1.60 & 96.49$\pm$1.68 & 85.34$\pm$7.43  \\
8  & Brick       & 94    & 18  & 3,570  & 3,682  & 84.74$\pm$5.07 & 67.43$\pm$8.04  & 87.20$\pm$4.04 & 90.51$\pm$4.16 & 92.44$\pm$2.03 & 76.15$\pm$11.94 \\
9  & Shadow      & 94    & 18  & 835    & 947    & 99.77$\pm$0.13 & 99.72$\pm$0.30  & 99.64$\pm$0.23 & 99.92$\pm$0.11 & 99.71$\pm$0.34 & 99.08$\pm$0.79  \\ \hline
\multicolumn{6}{c}{OA (\%)}                      & 62.15$\pm$1.68 & 69.50$\pm$3.22  & 86.97$\pm$3.27 & 96.17$\pm$0.80 & 91.78$\pm$1.30 & 81.07$\pm$3.65  \\
\multicolumn{6}{c}{AA (\%)}                      & 74.19$\pm$0.54 & 79.20$\pm$1.90  & 90.56$\pm$1.60 & 96.12$\pm$0.59 & 93.92$\pm$0.83 & 85.84$\pm$2.43  \\
\multicolumn{6}{c}{Kappa $\times$ 100}           & 54.01$\pm$1.56 & 61.77$\pm$3.37  & 83.10$\pm$4.02 & 94.90$\pm$1.05 & 89.19$\pm$1.61 & 75.83$\pm$4.30  \\ \hline
\end{tabular}}
\end{table*}
\begin{table*}[t!]
\centering
\caption{Class-wise and overall classification performance under the non-random spatial sampling setting designed to reduce same-class train-test spatial overlap. The reported values are presented as mean$\pm$standard deviation over repeated runs.}
\vspace{-1em}
\label{nonRandom}
\resizebox{0.85\linewidth}{!} {
\begin{tabular}{clcccccccccc}
\hline
No & Name        & Train & Val & Test   & Total  & SVM             & MLP             & 2D-CNN          & 3D-CNN          & ViT             & MorpMamba       \\ \hline
1  & Asphalt     & 94    & 18  & 6,519  & 6,631  & 61.04$\pm$12.66 & 64.43$\pm$9.44  & 50.30$\pm$18.10 & 46.27$\pm$16.32 & 46.63$\pm$14.61 & 49.11$\pm$15.48 \\
2  & Meadows     & 94    & 18  & 18,537 & 18,649 & 38.72$\pm$12.41 & 33.66$\pm$14.46 & 41.56$\pm$10.45 & 49.78$\pm$18.88 & 50.26$\pm$17.19 & 37.60$\pm$13.61 \\
3  & Gravel      & 94    & 18  & 1,987  & 2,099  & 29.85$\pm$12.74 & 36.51$\pm$22.81 & 34.83$\pm$22.72 & 43.99$\pm$20.93 & 31.36$\pm$17.46 & 30.97$\pm$20.54 \\
4  & Trees       & 94    & 18  & 2,952  & 3,064  & 76.88$\pm$19.95 & 80.39$\pm$13.05 & 62.17$\pm$18.31 & 82.36$\pm$11.12 & 57.62$\pm$21.13 & 73.17$\pm$19.35 \\
5  & Metal Sheet & 94    & 18  & 1,233  & 1,345  & 99.60$\pm$0.22  & 99.45$\pm$0.49  & 92.99$\pm$4.96  & 99.76$\pm$0.47  & 98.90$\pm$1.57  & 97.31$\pm$2.93  \\
6  & Bare Soil   & 94    & 18  & 4,917  & 5,029  & 45.81$\pm$11.61 & 48.65$\pm$12.35 & 38.41$\pm$6.55  & 47.89$\pm$13.47 & 39.51$\pm$11.01 & 42.46$\pm$14.88 \\
7  & Bitumen     & 94    & 18  & 1,218  & 1,330  & 66.79$\pm$19.21 & 74.52$\pm$17.45 & 72.00$\pm$16.20 & 89.92$\pm$5.38  & 64.78$\pm$20.57 & 77.78$\pm$15.34 \\
8  & Brick       & 94    & 18  & 3,570  & 3,682  & 63.36$\pm$12.02 & 57.38$\pm$24.16 & 53.63$\pm$17.66 & 56.55$\pm$20.72 & 56.75$\pm$17.02 & 64.94$\pm$15.13 \\
9  & Shadow      & 94    & 18  & 835    & 947    & 99.40$\pm$0.39  & 99.70$\pm$0.11  & 93.08$\pm$3.74  & 96.51$\pm$5.36  & 80.85$\pm$12.76 & 88.28$\pm$7.01  \\ \hline
\multicolumn{6}{c}{OA (\%)}                      & 51.25$\pm$5.43  & 50.15$\pm$6.91  & 48.16$\pm$5.96  & 55.20$\pm$9.21  & 51.07$\pm$7.42  & 48.45$\pm$9.26  \\
\multicolumn{6}{c}{AA (\%)}                      & 64.60$\pm$3.76  & 66.08$\pm$3.22  & 59.89$\pm$4.78  & 68.12$\pm$3.02  & 58.52$\pm$3.57  & 62.40$\pm$4.92  \\
\multicolumn{6}{c}{Kappa $\times$ 100}           & 41.84$\pm$5.09  & 41.27$\pm$6.44  & 38.06$\pm$6.45  & 45.96$\pm$8.64  & 40.54$\pm$6.81  & 39.15$\pm$9.06  \\ \hline
\end{tabular}}
\end{table*}

The experiments are conducted on the Pavia University hyperspectral dataset, which contains nine labeled land-cover classes \cite{dataURL}. Before patch extraction, PCA is applied to reduce the spectral dimensionality to $P=15$ components, and each labeled pixel is represented by a $9 \times 9 \times 15$ spatial-spectral patch.

For both sampling protocols, a class-balanced training set is constructed by selecting 94 samples per class, corresponding to approximately one-tenth of the smallest class, Shadow, which contains 947 labeled samples. In addition, 18 samples per class are used for validation and reserved only for model monitoring and selection, while all remaining labeled pixels are used for testing.

The evaluated classifiers include SVM \cite{melgani2004classification}, MLP \cite{plaza2008parallel}, 2D-CNN \cite{makantasis2015deep}, 3D-CNN \cite{hamida20183}, ViT \cite{dosovitskiy2020image}, and MorpMamba \cite{ahmad2025spatial}. The same training configuration is used under both sampling protocols so that performance differences are mainly attributed to the sampling strategy. All experiments are repeated over 10 independent runs, and the results are reported as mean$\pm$standard deviation using Overall Accuracy (OA), Average Accuracy (AA), Kappa coefficient, and class-wise accuracy.

\section{Results and Discussion}
\vspace{-1em}
Tables~\ref{Random} and~\ref{nonRandom} present the
classification results under random and non-random patch-based
sampling, respectively. Under random sampling, deep spectral-spatial models clearly outperform traditional classifiers. The 3D-CNN achieves the highest OA of 96.17\%, followed by ViT, 2D-CNN, and MorpMamba with 91.78\%, 86.97\%, and 81.07\%, respectively,
whereas SVM and MLP obtain 62.15\% and 69.50\%. A similar trend
is observed for AA and Kappa, indicating that deep patch-based
models benefit strongly when training and testing samples are
randomly distributed within the same image.

This advantage decreases substantially under the non-random spatial setting. As summarized in Table~\ref{tab:oa_drop}, all models experience an OA reduction, with the largest drops observed for 3D-CNN, ViT, and 2D-CNN at 40.97, 40.71, and 38.81 percentage points, respectively. MorpMamba decreases by 32.62 percentage points, while SVM and MLP show smaller drops of 10.90 and 19.35 percentage points. These results indicate that models explicitly exploiting spatial neighborhoods are more sensitive to the sampling protocol.

The class-wise results show a similar pattern. For 3D-CNN, accuracy decreases from 94.28\% to 46.27\% for Asphalt, from 97.63\% to 49.78\% for Meadows, from 91.59\% to 43.99\% for Gravel, and from 96.38\% to 47.89\% for Bare Soil. These large reductions mainly occur in spatially variable classes. In contrast, compact and spectrally distinctive classes, such as Metal Sheet and Shadow, remain comparatively easier to classify under non-random sampling.

Overall, Tables~\ref{Random} and~\ref{nonRandom} indicate that the
high accuracy obtained under random patch-based evaluation is partly influenced by same-class train-test spatial dependency. Neighboring patches may share pixels or highly similar local structures, allowing information from training regions to be indirectly reflected in the test set. Random sampling may therefore produce optimistic performance estimates, whereas non-random spatial sampling provides a stricter assessment of model generalization.

\begin{table}[t!]
\centering
\caption{Overall accuracy degradation, in percentage points, from random sampling to non-random spatial sampling.}
\vspace{-1em}
\label{tab:oa_drop}
\resizebox{0.9\linewidth}{!}{
\begin{tabular}{lccc}
\hline
Model & Random OA (\%) & Non-Random OA (\%) & OA Drop (pp) \\ \hline
SVM       & 62.15 & 51.25 & 10.90 \\
MLP       & 69.50 & 50.15 & 19.35 \\
2D-CNN    & 86.97 & 48.16 & 38.81 \\
3D-CNN    & 96.17 & 55.20 & 40.97 \\
ViT       & 91.78 & 51.07 & 40.71 \\
MorpMamba & 81.07 & 48.45 & 32.62 \\ \hline
\end{tabular}}
\end{table}

\subsection{Effect of Patch Size on Same-Class Spatial Leakage}
\vspace{-0.5em}
Table~\ref{tab:patch_overlap_ratio} shows the effect of patch size on the same-class $OP$ and $AOR$, computed using Eqs.~\eqref{eq:overlap_percentage} and~\eqref{eq:average_overlap_ratio}, respectively. Under random sampling, $OP$ increases from 23.28\% with a $5 \times 5$ patch to 77.02\% with a $19 \times 19$ patch, showing that larger patches substantially increase the portion of testing regions covered by training patches. $AOR$ follows the same trend, increasing from 47.29\% to 79.91\%, which indicates that larger patches also increase the severity of overlap among the affected testing patches.
\begin{table}[t!]
\centering
\caption{Effect of patch size on same-class overlap percentage and average overlap ratio under random and non-random spatial sampling settings. The values are computed from the training and testing patch locations, not from model predictions. Only mean values are reported.}
\vspace{-1em}
\label{tab:patch_overlap_ratio}
\resizebox{0.9\linewidth}{!}{
\begin{tabular}{ccc|cc}
\hline
\multirow{2}{*}{Patch Size} & \multicolumn{2}{c|}{Overlap Percentage (\%)} & \multicolumn{2}{c}{Average Overlap Ratio (\%)} \\ \cline{2-5}
& Random & Non-Random & Random & Non-Random \\ \hline
$5 \times 5$   & 23.28 & 0.83 & 47.29 & 45.26 \\
$7 \times 7$   & 35.03 & 1.28 & 52.98 & 43.79 \\
$9 \times 9$   & 45.29 & 1.75 & 58.45 & 42.26 \\
$11 \times 11$ & 54.04 & 2.26 & 63.61 & 40.50 \\
$13 \times 13$ & 61.48 & 2.80 & 68.31 & 38.98 \\
$15 \times 15$ & 67.73 & 3.39 & 72.70 & 38.86 \\
$17 \times 17$ & 72.86 & 4.01 & 76.60 & 39.13 \\
$19 \times 19$ & 77.02 & 4.66 & 79.91 & 39.42 \\ \hline
\end{tabular}}
\end{table}
In contrast, the non-random spatial setting maintains a much lower overlap percentage, increasing only from 0.83\% to 4.66\% as the patch size grows from $5 \times 5$ to $19 \times 19$. This confirms that the non-random sampling strategy greatly reduces the global amount of same-class train-test overlap. However, $AOR$ remains relatively high, decreasing from 45.26\% to around 39\% for larger patch sizes. This is because Eq.~\eqref{eq:average_overlap_ratio} is computed only over testing patches with non-zero overlap; thus, although few testing patches are affected, the remaining overlap cases may still contain considerable local overlap.

Overall, patch size directly controls both the extent and severity of spatial leakage. $OP$ measures the global amount of leakage, whereas $AOR$ provides complementary information about the severity of the affected cases. This distinction is important in leakage-reduced settings, where the number of affected testing patches may be small even when their local overlap remains high.

\section{Conclusion}
\vspace{-1em}
This paper investigated same-class train-test spatial overlap in patch-based hyperspectral image classification using two measures: $OP$ and $AOR$. The results showed that random patch-based sampling can substantially inflate classification performance, especially for deep models that exploit spatial context. Under non-random spatial sampling, 3D-CNN, ViT, and 2D-CNN showed large OA drops of 40.97, 40.71, and 38.81 percentage points, respectively, while SVM and MLP showed smaller reductions. The patch-size analysis further confirmed that larger patches increase spatial leakage, with the random-sampling overlap percentage rising from 23.28\% for $5 \times 5$ patches to 77.02\% for $19 \times 19$ patches.

Future work will extend this analysis to additional hyperspectral datasets and different training ratios to further assess the generality of the observed leakage effects. More rigorous spatial separation strategies will also be investigated to reduce train-test dependency while preserving sufficient labeled samples for training. In addition, future studies may examine how leakage affects newer deep architectures and whether model design choices can improve robustness under leakage-reduced evaluation settings.

\patchcmd{\thebibliography}{\settowidth}{\setlength{\itemsep}{0pt}\setlength{\parskip}{0pt plus 1.5ex}\settowidth}{}{}
\vspace{-0.5em}
\bibliographystyle{IEEEbib}
\balance
\bibliography{00_Main_Doc}

@article{melgani2004classification,
  title={Classification of hyperspectral remote sensing images with support vector machines},
  author={Melgani, Farid and Bruzzone, Lorenzo},
  journal={IEEE Transactions on geoscience and remote sensing},
  volume={42},
  number={8},
  pages={1778--1790},
  year={2004},
  publisher={IEEE}
}

@inproceedings{makantasis2015deep,
  title={{Deep supervised learning for hyperspectral data classification through convolutional neural networks}},
  author={Makantasis, Konstantinos and Karantzalos, Konstantinos and Doulamis, Anastasios and Doulamis, Nikolaos},
  booktitle={2015 IEEE international geoscience and remote sensing symposium (IGARSS)},
  pages={4959--4962},
  year={2015},
  organization={IEEE}
}

@article{Alkhatib26042026,
author = {Mohammed Q. Alkhatib},
title = {{ConvViTMamba: efficient multiscale convolution, Transformer, and Mamba-based sequence modelling for hyperspectral image classification}},
journal = {International Journal of Remote Sensing},
volume = {0},
number = {0},
pages = {1--40},
year = {2026},
publisher = {Taylor \& Francis},
doi = {10.1080/01431161.2026.2663567},
}

@incollection{plaza2008parallel,
  title={Parallel classification of hyperspectral images using neural networks},
  author={Plaza, Javier and Plaza, Antonio and P{\'e}rez, Rosa and Mart{\'\i}nez, Pablo},
  booktitle={Computational intelligence for remote sensing},
  pages={193--216},
  year={2008},
  publisher={Springer}
}

@misc{dataURL,
  title = {Hyperspectral Remote Sensing Scenes},
  howpublished = {\url{https://www.ehu.eus/ccwintco/index.php/Hyperspectral_Remote_Sensing_Scenes}},
  note = {Accessed: 31-May-2026}
}

@article{dosovitskiy2020image,
  title={An image is worth 16x16 words: Transformers for image recognition at scale},
  author={Dosovitskiy, Alexey and Beyer, Lucas and Kolesnikov, Alexander and Weissenborn, Dirk and Zhai, Xiaohua and Unterthiner, Thomas and Dehghani, Mostafa and Minderer, Matthias and Heigold, Georg and Gelly, Sylvain and others},
  journal={arXiv preprint arXiv:2010.11929},
  year={2020}
}

@article{zhan2026rethinking,
  title={Rethinking Data Leakage in Patch-Based Hyperspectral Image Classification with Traditional Deep Networks},
  author={Zhan, Kaizhe and Huo, Zijie and Yin, Zhijian and Yang, Zhen},
  journal={Remote Sensing},
  volume={18},
  number={7},
  pages={979},
  year={2026},
  publisher={MDPI}
}

@article{feng2023information,
  title={Information leakage in deep learning-based hyperspectral image classification: A survey},
  author={Feng, Hao and Wang, Yongcheng and Li, Zheng and Zhang, Ning and Zhang, Yuxi and Gao, Yunxiao},
  journal={Remote Sensing},
  volume={15},
  number={15},
  pages={3793},
  year={2023},
  publisher={MDPI}
}

@article{chang2023iterative,
  title={Iterative random training sampling convolutional neural network for hyperspectral image classification},
  author={Chang, Chein-I and Liang, Chia-Chen and Hu, Peter Fuming},
  journal={IEEE Transactions on Geoscience and Remote Sensing},
  volume={61},
  pages={1--26},
  year={2023},
  publisher={IEEE}
}

@article{ahmad2022disjoint,
  title={A disjoint samples-based 3D-CNN with active transfer learning for hyperspectral image classification},
  author={Ahmad, Muhammad and Ghous, Usman and Hong, Danfeng and Khan, Adil Mehmood and Yao, Jing and Wang, Shaohua and Chanussot, Jocelyn},
  journal={IEEE Transactions on Geoscience and Remote Sensing},
  volume={60},
  pages={1--16},
  year={2022},
  publisher={IEEE}
}

@article{ahmad2025spatial,
  title={Spatial--spectral morphological mamba for hyperspectral image classification},
  author={Ahmad, Muhammad and Butt, Muhammad Hassaan Farooq and Khan, Adil Mehmood and Mazzara, Manuel and Distefano, Salvatore and Usama, Muhammad and Roy, Swalpa Kumar and Chanussot, Jocelyn and Hong, Danfeng},
  journal={Neurocomputing},
  volume={636},
  pages={129995},
  year={2025},
  publisher={Elsevier},
  doi = {10.1016/j.neucom.2025.129995}
}

@inproceedings{alkhatib2024hsiformer,
  title={{HSIFormer: An Efficient Vision Transformer Framework for Enhanced Hyperspectral Image Classification Using Local Window Attention}},
  author={Alkhatib, Mohammed Q and Jamali, Ali},
  booktitle={2024 14th Workshop on Hyperspectral Imaging and Signal Processing: Evolution in Remote Sensing (WHISPERS)},
  pages={1--5},
  year={2024},
  organization={IEEE}
}

@article{roy2019hybridsn,
  title={{HybridSN: Exploring 3-D--2-D CNN feature hierarchy for hyperspectral image classification}},
  author={Roy, Swalpa Kumar and Krishna, Gopal and Dubey, Shiv Ram and Chaudhuri, Bidyut B},
  journal={IEEE Geoscience and Remote Sensing Letters},
  volume={17},
  number={2},
  pages={277--281},
  year={2019},
  publisher={IEEE}
}

@article{hamida20183,
  title={{3-D deep learning approach for remote sensing image classification}},
  author={Hamida, Amina Ben and Benoit, Alexandre and Lambert, Patrick and Amar, Chokri Ben},
  journal={IEEE Transactions on geoscience and remote sensing},
  volume={56},
  number={8},
  pages={4420--4434},
  year={2018},
  publisher={IEEE}
}

@article{hu2015deep,
  title={{Deep convolutional neural networks for hyperspectral image classification}},
  author={Hu, Wei and Huang, Yangyu and Wei, Li and Zhang, Fan and Li, Hengchao},
  journal={Journal of Sensors},
  volume={2015},
  pages={1--12},
  year={2015},
  publisher={Hindawi Limited}
}

\end{document}